\pdfoutput=1
\documentclass[11pt]{article}

\usepackage[]{ACL2023}

\usepackage{times}
\usepackage{latexsym}
\usepackage{subfigure}
\usepackage{graphicx}
\usepackage{amsthm}
\usepackage{amsmath}
\usepackage{bm}
\usepackage{bbding}

\usepackage[T1]{fontenc}
\usepackage[utf8]{inputenc}

\usepackage{microtype}

\usepackage{inconsolata}

\usepackage{amsfonts,amssymb}
\usepackage{float}
\usepackage{algorithm}
\usepackage{algorithmic}

\usepackage{xcolor}

\usepackage{multicol}
\usepackage{multirow}

\title{LLM-based Medical Assistant Personalization with Short- and Long-Term Memory Coordination}

\author{Kai Zhang$^{1}$, Yangyang Kang$^{2}$, Fubang Zhao$^{2}$, Xiaozhong Liu$^{1}$\Thanks{Corresponding Author} \\ $^{1}$Worcester Polytechnic Institute, Worcester, USA \\ $^{2}$Alibaba Group, Hangzhou, China \\
\texttt{\{kzhang8, xliu14\}@wpi.edu, \{yangyang.kangyy, fubang.zfb\}@alibaba-inc.com} \\}

\begin{document}
\maketitle

\begin{abstract}

Large Language Models (LLMs), such as GPT3.5, have exhibited remarkable proficiency in comprehending and generating natural language. On the other hand, medical assistants hold the potential to offer substantial benefits for individuals. However, the exploration of LLM-based personalized medical assistant remains relatively scarce.  Typically, patients converse differently based on their background and preferences which necessitates the task of enhancing user-oriented medical assistant. While one can fully train an LLM for this objective, the resource consumption is unaffordable. Prior research has explored memory-based methods to enhance the response with aware of previous mistakes for new queries during a dialogue session. We contend that a mere memory module is inadequate and fully training an LLM can be excessively costly. In this study, we propose a novel computational bionic memory mechanism, equipped with a parameter-efficient fine-tuning (PEFT) schema, to personalize medical assistants. To encourage further research into this area, we are releasing a new conversation dataset generated based on an open-source medical corpus and our implementation code\footnote{https://github.com/MatthewKKai/MaLP}.

\end{abstract}

\section{Introduction}

\begin{figure}[t]
    \centering
    \includegraphics[scale=0.4]{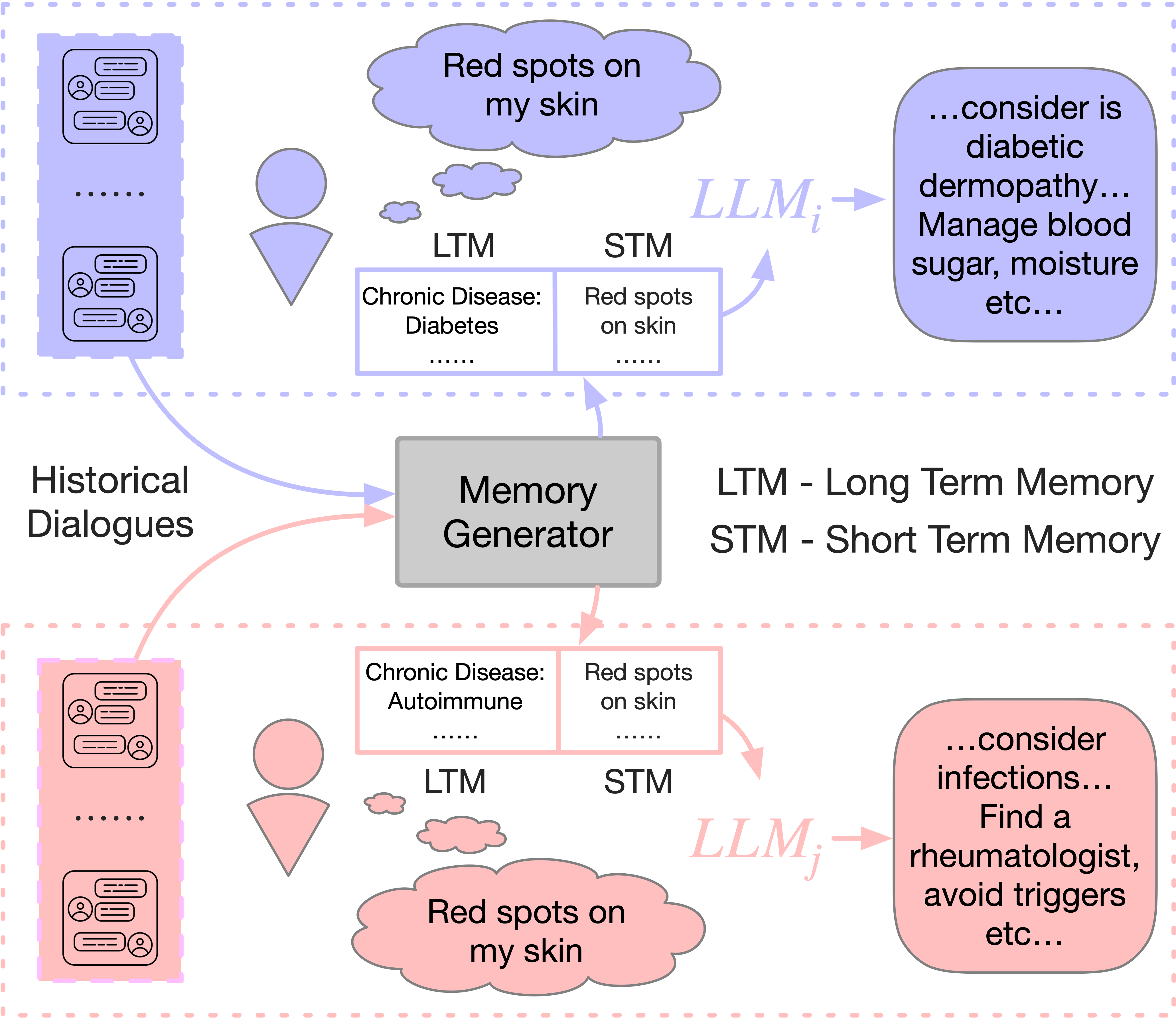}
    \caption{Personalized responses for different users in terms of the same query.}
    \label{fig:toy_example}
    \vspace{-8pt}
\end{figure}

The potential of large language models to understand and generate natural language is undeniable \citep{brown2020language, chowdhery2022palm, touvron2023llama}, while there is an untapped opportunity to explore how LLMs could be customised to provide personalized medical advice with patients, allowing them to receive tailored responses that best suit their individual needs \citep{bender-koller-2020-climbing}. For example, as depicted in Figure \ref{fig:toy_example}, medical practitioners can discern vital patient information through ongoing diagnostic conversations. Consequently, responses to identical queries may differ based on individual patient nuances, highlighting the imperative need for personalized medical assistants leveraging LLM. Efforts have been made to obtain proper prompts for steering LLMs to enhance outputs. For example, by memorizing previous mistakes and user feedback, given a new query, a similarity-based retriever can be leveraged to preemptively recognize and rectify LLM errors\citep{dalvi2022towards, madaan2022memory, lewis2020retrieval}. However, this paradigm poses us two challenges: \textbf{Firstly}, most existing memory designs are dictionary-based \cite{madaan2022memory, lewis2020retrieval} (i.e. key-value form where key is the previous mistake, value is the corresponding user-feedback) which can be inflexible and rely heavily on the power of retriever. \textbf{Secondly}, such paradigm, without retraining, can barely provide users with personalized and engaging experience. For instance, a diabetes patient who prefers concise and straightforward medical advice won't expect detailed glucose test explanations from a doctor while others who prefer fully elaborated responses may want to know as much as possible about the disease (e.g., causes etc.). To this end, how to process patient-relevant information properly and being aware of their preference can be crucial for enhancing patients' experience and remains understudied. In this paper, we propose a novel memory mechanism along with a PEFT schema to enhance LLM-based medical assistant personalization.

Dictionary-based memory is not pliable due to its intricate structure and thus efforts can only be made in strengthening retrievers. Despite the improvements made by retrievers like semantic-similarity based and distance-closest based\cite{madaan2022memory}, we argue that the memory structure should be ameliorated to accommodate diverse information. Unfortunately, rare efforts have been made to address this challenge. Neuroscientists have revealed that real-world memory mechanism works in two processes, one can be characterized as automatic and unconscious while the other one is effortful and analytical\citep{kahneman2011thinking}. For example, acquiring a new skill initially can be active and analytical, but as one's proficiency increases, it becomes more intuitive. This is referred as Dual-process theory and correspondingly, memory can be defined as three types: working memory, short-term memory (STM) and long-term memory (LTM). Working memory is responsible for filtering and buffering information, STM holds knowledge for a short period, while LTM stores knowledge for a longer duration \citep{roediger1995creating}. Drawing inspirations from this, we propose a novel \textbf{D}ual-\textbf{P}rocess enhanced \textbf{M}emory (\textbf{DPeM}) mechanism of which three types of memory cooperate smoothly under the guidance of dual-process schema and thus can provide LLM more useful knowledge from both user-specific and common-sense aspects. \par
Furthermore, existing works in personalized LLMs primarily focus on designing comprehensive prompts that enable LLMs to generate contextually relevant responses aligned with user-specific dialogue preferences (e.g., preferring concise responses) \citep{wang2023chain, wu2023tidybot, wang2019repairing}. However, these methods often yield inferior performance compared to fine-tuning approaches and are susceptible to the exact formatting of the prompts (e.g., wording and ordering) \citep{liu2022few}. Another approach proposed by \citet{salemi2023lamp} involves incorporating user profiles during the pretraining stage, enabling LLMs to possess user-specific knowledge for downstream tasks. Nevertheless, training fully personalized LLMs for individuals can be economically unviable. To this end, we embrace the utilization of PEFT which focuses on updating a small subset of parameters, ensuring that the trained LLM achieves promising performance on new tasks while minimizing computational costs, to develop user-oriented LLMs with reduced time and resource consumption. \par
% PEFT can enable LLM to learn to respond to users while being aware of their specific needs in a computationally and resource-efficient way. \par
In tandem, we propose a novel memory mechanism inspired by neuroscience, and along with a PEFT training strategy to achieve LLM-based medical assistant personalization. The key contributions of our work are as follows:\par
$\bullet$ We propose a novel DPeM mechanism that closely resembles real-world memory processes which lead to a relatively 7\% improvement against existing memory structure.\par
$\bullet$ We propose MaLP, a unified frame based on DPeM and PEFT which promotes the response's quality by catering to user-specific needs. \par
$\bullet$ We introduce a new medical dialogue dataset that incorporates user preferences and historical records. This dataset offers a unique perspective to explore personalized medical assistants. \par

\section{Methodology}
\subsection{Preliminary Definition}
Before going further, we would like to give our preliminary definitions first. Given multi-round dialogues between two characters (e.g., patient and doctor) which is denoted as $\mathcal{D} = \{d_{0}, ..., d_{n}\}$ where $n$ is the number of rounds, our task here is to learn and memorize the knowledge from $\mathcal{D}$ to form a memory $\mathcal{M}$ and fine-tune a large language model (LLM) $\Phi$ to produce personalized response $y$ in terms of a new query $x$ from the same user with respect to (w.r.t) $\mathcal{D}$ and $\mathcal{M}$. \par

\subsection{Medical Knowledge Adaptation}
To help the LLM provide better responses, we propose to first incorporate more medical knowledge via a domain adapter\citep{zhang2023llama}. The adapter architecture consists of a down-projection layer, a non-linearity function (e.g., ReLU\citep{agarap2018deep}), and an up-projection (e.g., a fully connected network). Note that all parameters, except those pertaining to the domain adapter, remain frozen. However, directly using such domain adapter will lead to the catastrophic forgetting problem\citep{gururangan2020don}. This phenomenon entails the risk that the LLM may lose its inherent capabilities after training on domain-specific knowledge using an adapter. To solve this, we propose leveraging a sample loss, which gauges the output disparity before and after the knowledge adaptation process for the same query. Given a medical text with K masked tokens, the knowledge loss can be $\mathcal{L}_{K} = -\frac{1}{K}\sum_{i=1}^{M}log\ p(m_{i})$ where $p(m_{i})$ is the probability of generating $m_{i}$, and the sample loss can be defined as $\mathcal{L}_{S} = ||V_{o}, V_{k}||_{2}^{2}$, where the $V_{o}$ is the vector representation of tokens from the original layer and $V_{k}$ is the vector representation of tokens from the layer that installed the adapter. The overall knowledge adaptation can be trained by simply adding those two losses. We use the trained LLM with external medical knowledge as the base LLM in the following steps.

\subsection{DPeM Mechanism}
\begin{figure}[ht]
    \centering
    \includegraphics[width=7.7cm]{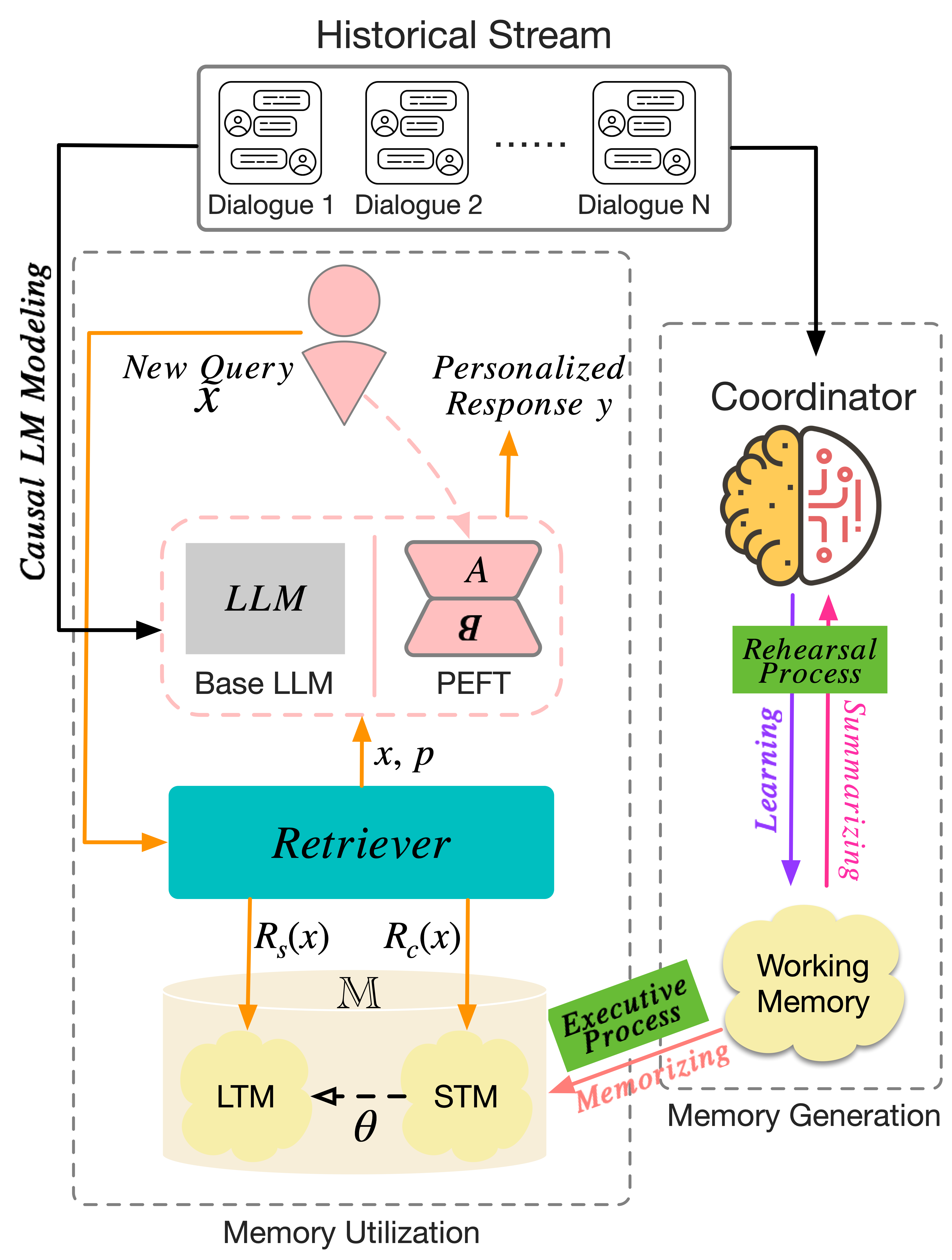}
    \caption{Overview of MaLP: the user's historical dialogues will firstly be passed to a coordinator $\mathcal{C}$ and a trainable LLM equipped with PEFT iteratively for memory generation and causal language modeling, respectively. Then the memory generation module will form a memory using DPeM mechanism where dual-process is denoted in green box along with three steps denoted in colored lines separately. After iterations completed, a new query by the user will be passed to a retriever for corresponding memory lookup and then the fine-tuned LLM will produce the personalized response in terms of the retrieved knowledge and historical dialogues.}
    \label{fig:overview}
    \vspace{-8pt}
\end{figure}

The key novelty that differentiates our work with previous efforts is that we turns into excavating the improvements regarding the intricate memory structure instead of studying solely on retriever. Drawing inspirations from \citet{kahneman2011thinking}, we aim to design a memory mechanism that closely resembles real-world memory processes. To achieve this, we propose a dual-process (Rehearsal Process, Executive Process) enhanced procedure which consists of three steps - Learning, Summarizing and Memorizing. The Rehearsal Process involves learning information from $\mathcal{D}$, which is then stored in working memory. The working memory is refreshed iteratively based on the dialogue's content of current iteration—this is the summarizing step. The two-step rehearsal process is facilitated by a coordinator with powerful natural language understanding abilities. Next, the information stored in working memory is evaluated to determine whether it needs to be stored in Short-Term Memory (STM) or Long-Term Memory (LTM) based on the frequency of access by the Executive Process. This dual-process is illustrated in Figure \ref{fig:overview} by the green-box and the three-colored lines. The detailed memory structure and working flows of DPeM are depicted in the following sections. 

\subsubsection{Memory $\mathcal{M}$:} 
As prescribed, $\mathcal{M}$ does not consist of a single type of memory. Instead, it comprises different types of memory that store and access information in their own way, while working together for better knowledge management through dual-process. Intuitively, there will be a vast amount of information that needs to be registered when learning something new and not all those information will be stored directly and entirely into memory. Instead, a working memory acts as a buffer memory to register and filter information so that only the relevant information enters STM, while the rest is dropped. Short-term memory refers to a limited space that holds a small amount of knowledge in an active, quickly accessible state. Long-term memory stores knowledge transited from STM for a longer period. Being aware of the differences in information storage and access, our memory mechanism enhances LLM by incorporating knowledge from both user-specific and common-sense perspectives. All the three types of memory are in the form of a growing table which support different operations but work collaboratively as shown in Table \ref{tab:memory_comparison}.
\subsubsection{Rehearsal Process} 
Rehearsal refers to the process of obtaining and refreshing information so that the relevant information can be filtered and stored as knowledge into memory. \textbf{Learning} aims at gathering information from the dialogue's content of the current iteration.  Inspired by how humans take notes when learning something new, we leverage $\mathcal{C}$, which takes all the necessary notes for each iteration's dialogue. Formally, by passing $d_{i}\in \mathcal{D}$ to $\mathcal{C}$ iteratively, we obtain some notes denoted as $nts = \mathcal{C}(d_{i})$ which will be recorded into working memory. Typically, not all the notes will be practical which makes the summarizing step indispensable within the DPeM mechanism. \textbf{Summarizing} steps further by purifying the relevant notes from $nts$ and pass them into STM. Specifically, the coordinator will determine if the $nt\in nts$ is relevant or not and store the useful $nt^{+}$ as knowledge item by item and the stored knowledge is delivered to STM and is denoted as $K = [k_{0}, ..., k_{m}]$ where $k_{i} = nt_{i}^{+}$.

\begin{table}[t]
    \centering
    \small
    \begin{tabular}{cccc}
        \hline
        Type & Refresh & Storage & Sup. Lookup?\\
        \hline
        $\mathcal{M}_{working}$ & Each Iteration & Limited & \XSolidBrush \\
        $\mathcal{M}_{STM}$ & Certain Rounds & Limited & \Checkmark \\
        $\mathcal{M}_{LTM}$ & Never & Unlimited & \Checkmark  \\
        \hline
    \end{tabular}
    \caption{Comparison among three types of memory.}
    \label{tab:memory_comparison}
\end{table}

\subsubsection{Executive Process} 
Executive Process mainly focuses on and \textbf{memorizing} the knowledge produced from \textit{rehearsal process}. The main objective of DPeM is to process and store information based on its importance level and user-specific needs, an aspect that previous works have paid little attention to. Specifically, the filtered knowledge $k$ will firstly be categorized as two types: Common-sense Knowledge, User-Specific Knowledge and then be converted into the STM in the form of key (type) - value ($k_{i}$) pair. As the learning iteration progresses, a flag table $ft$ is used to keep track of the frequency of appearance for each $k_{i}$. When the frequency reaches a predetermined threshold $\theta$, the $k_{i}$ is transferred to LTM. Notably, STM is refreshed periodically after certain rounds (working memory is refreshed after each iteration) while LTM typically only accepts new $k_{i}$ entries. The final memory structure consists of three parts: Working Memory, STM, and LTM. Working Memory serves as a buffer for storing newly detected information, STM stores relevant and recent knowledge, and LTM provides longer-term access to frequently visited knowledge from STM. Through the collaboration of these three types of memory, along with the dual-process approach, DPeM provides a more powerful memory system to further support personalized LLM.
\subsection{MaLP Frame}
\subsubsection{Memory Generation} Memory can provide latent knowledge from personal historical stream which can further be neglected as prompts to assist LLM for producing desired responses regarding new queries. Attributed to our proposed DPeM mechanism, the memory generation module can produce a well-organized memory which can support different storage and lookup operations in terms of information features as can be seen in Fig. \ref{fig:overview}. Given dialogues $\mathcal{D}$, the memory formation can be described as follows:
\begin{equation}
\begin{split}
    \mathcal{M}_{working} &= \{nt_{0}, ..., nt_{i}, ...\}, \\
    \mathcal{M}_{STM} &= \{..., k\_type: k_{j}, ...\}, \\
    \mathcal{M}_{LTM} &= \{...,\ k\_type: k_{f},\ ...\}, \\
    \mathcal{M} &= [\mathcal{M}_{working}, \mathcal{M}_{STM}, \mathcal{M}_{LTM}]\\
\end{split}
    \label{memory_generation}
\end{equation}
 where $nt_{i} = \mathcal{C}(d_{i})$, $k_{j} = nt_{i}^{+}$, $k_{f}$ denotes frequently visited $k_{j}$ from $\mathcal{M}_{STM}$. The comparison among these three types of memory can be seen in the Table \ref{tab:memory_comparison}.
\subsubsection{Memory Utilization} However, relying solely on memory for achieving personalized LLMs still poses challenges, as the quality of generated responses ultimately depends on the understanding and generation ability of the LLM, even with memory-augmented prompts and pre-injected knowledge. Therefore, fine-tuning the LLM to cater to user-specific needs naturally becomes an option for enhancing LLM personalization. However, traditional fine-tuning approaches often demand significant computational and data resources, whereas our aim is to optimize the LLM's response generation in a user-friendly manner by leveraging previous dialogues. In this regard, PEFT methods \citep{li-liang-2021-prefix, liu-etal-2022-p, liu2022ptuning} offer a solution by achieving this objective with low resource consumption.  \par
To tune the base LLM (e.g., LLaMA) with user's previous dialogues and enable it to generate user-favorable responses, we employ the Low-Rank Adaption (LoRA) technique \citep{hu2021lora}. With LoRA, we update a given pre-trained weight matrix $W_{\Phi}\in \mathbb{R}^{d\times k}$ of LLM by incorporating a low-rank decomposition $W_{\Phi}+\Delta W = W_{\Phi}+BA$. Here, $B\in \mathbb{R}^{d\times r}$, $A\in \mathbb{R}^{r\times k}$, and the rank $r\ll \min(d, k)$. During the fine-tuning process, we randomly select a set number of layers to implement LoRA, where $A$ and $B$ are trainable while $W_{\Phi}$ remains frozen. This allows us to target user preferences and adapt the LLM accordingly. \par
Once all iterations are completed, we acquire a LoRA-tuned LLM along with a latent memory that caters to user-specific needs. Typically, when a new query $x$ is received, the responding process is presented as:
\begin{equation}
    x\rightarrow \Phi\rightarrow y
\end{equation}
however, by using MaLP, the process is refined as:
\begin{equation}
    \begin{split}
        &p = Retriever(x), \\
        &x, p\rightarrow \hat{\Phi}\rightarrow y
    \end{split}
\end{equation}
where $p$ is the prompt retrieved from $\mathcal{M}$, $Retriever$ is a function that can retrieve knowledge from $\mathcal{M}$ in terms of query $x$ and $\hat{\Phi}$ is the LoRA-tuned LLM. The utilization process is denoted in orange lines as can be seen in Figure \ref{fig:overview}. 
\subsubsection{Components}In the MaLP framework, several key components actively engage in memory generation and utilization to ensure efficient collaborations: \\
\textbf{Coordinator $\mathcal{C}$:} $\mathcal{C}$ plays a pivotal role in the learning and summarizing stage which involves deriving information from dialogue contexts and purifying knowledge from learned information \citep{xu2023baize}. Thus we resort to a powerful tool (e.g., ChatGPT) that is capable of understanding the long dialogue and performing summarization and judgement\citep{xu2023baize}.\\ 
\textbf{Retriever $\mathcal{R}$:} Retrieval is the process by which the retriever accesses stored knowledge. However, since the memorized knowledge differs between STM and LTM, their retrieval processes also differ. STM is retrieved in the order in which it is stored, while LTM is retrieved through association(e.g., recalling a past mistake by recognizing its similarity\citep{kahneman2011thinking}). To address this, we have designed two retrievers: a closest-match retriever, $\mathcal{R}{c}$, for STM retrieval, and a semantic-match retriever, $\mathcal{R}{s}$, for LTM retrieval. $\mathcal{R}_{c}$ aims to find the knowledge stored in STM that is closest to the query in terms of Levenshtein distance, which indicates the minimum number of deletions, insertions, or substitutions required to transform string s into string t (e.g., $lev('test', 'tent')=1$ since only one step $'s'->'n'$ is needed). However, since the retrieval process for LTM is fast and unconscious, we have chosen to train an encoder to obtain semantic embeddings and retrieve knowledge in LTM based on cosine similarity \citep{madaan2022memory}.

\section{Data}
\subsection{Data Construction}

\begin{figure}
    \centering
    \includegraphics[scale=0.6]{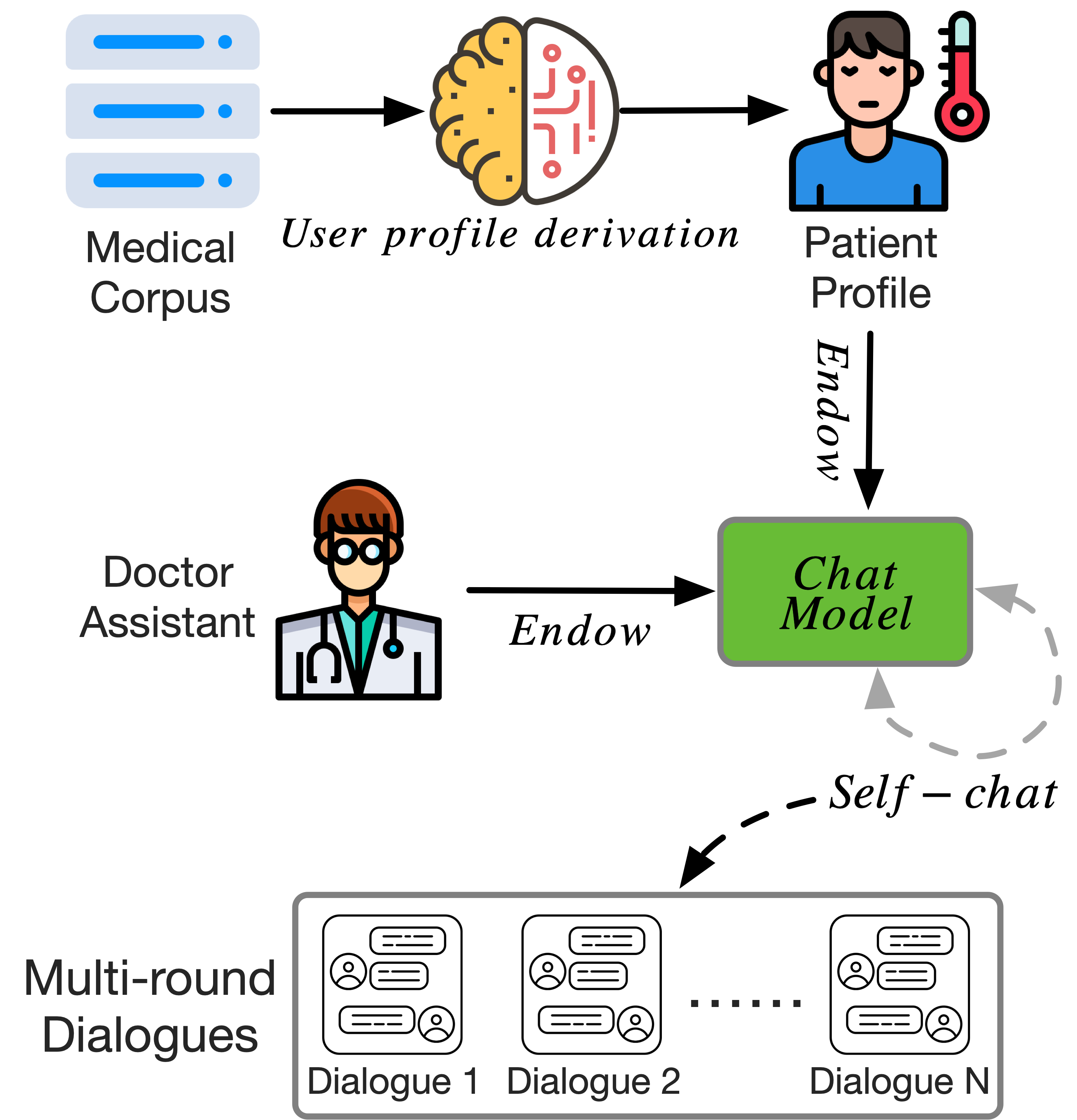}
    \caption{Details of data collection process: we first derive patient's profile from public medical corpus and then endow the patient's profile to a powerful chat model. Assistant role (e.g., doctor) will be simulated independently using the same chat model and thus we could collect the historical dialogues via \textit{self-chat} between these two roles.}
    \label{fig:data_collection}
\end{figure}

Existing dialogue datasets often lack awareness of the importance of penalization, while recent works like \citet{xu2023baize} have explored the capability of LLMs to generate high-quality chat corpora. In light of this, we propose injecting user profiles into the dialogue generation process using \textit{self-chat} simulations within real-world conversational scenarios. Specifically, we focus on medical scenarios as they typically involve dialogues between patients and doctors, encompassing a wealth of common-sense information (e.g., Tylenol can alleviate fever) and personal details (e.g., chronic diseases, dialogue preferences). These scenarios allow us to emphasize the significance of memory and personalization, respectively. To obtain personalized dialogues, one straightforward method is to incorporate user's profile into a language model prompt. We first derive the patient's profile including personal information, symptoms and dialogue preference from the publicly available medical corpus\footnote{https://github.com/UCSD-AI4H/Medical-Dialogue-System} \citep{chen2020meddiag} and then follow \citet{xu2023baize}'s work using \textit{self-chat} to guide powerful chat models (e.g., ChatGPT) simulate high-quality dialogues. The difference is that we endow the patient's and doctor's profile to the chat model at the beginning of conversation simulation. Further, we prompt the powerful chat model to produce follow-up dialogues related to the same symptom, new symptoms etc. to obtain historical information. The whole construction pipeline and detailed statistics can be seen in Figure \ref{fig:data_collection} and Appendix \ref{sec:appendix_a}, respectively.

\subsection{Safety and Evaluation}
\textbf{Safety} Unlike most dialogue data generated by chatting with human, our dataset does not rely on human feedback to suppress unwanted content (e.g., incorrect medicine suggestions) and instead we resort to providing explicit prompts that can steer the generation behaviors. While we have tested the default prompts, it can still be risky to have the guidance missed by changing the prompts. \\
\textbf{Evaluation} To further assess our dataset, two master students with medical background are hired for evaluating the quality of the dataset by identifying any dirty content and safety issues on 100 random samples\footnote{Quality scoring is depicted in the Appendix \ref{sec:appendix_b}}. The average quality score was found to be 5.27, and the safety ratio, indicating the proportion of instances without safety issues, was 94\%.

\section{Experiments}
\subsection{Setup}
For the medical knowledge injection, we use the open-source datasets from HealthCareMagic-100k and iCliniq\footnote{https://github.com/Kent0n-Li/ChatDoctor} \citep{yunxiang2023chatdoctor} and set the learning rate as 1e-4, batch size as 20, and weight decay as 0.05 for training. To train our MaLP, we leverage the AdamW optimizer\citep{loshchilov2018decoupled} with a learning rate of 5e-5 and also a linear warm-up scheduler initialized with 10\% of the total training steps as warm-up steps and a weight decay of 1e-4 to avoid over-fitting. The LoRA's rank of update matrices is set as 8 and the scaling factor alpha is 32. To accommodate the task of requirements, we set the maximum length of input and output to 1024 and 2048 tokens, respectively. All implementations are conducted with Pytorch\citep{paszke2017automatic}, PEFT\citep{peft} and Transformers\citep{wolf-etal-2020-transformers} on a computation node configured with a 256G CPU and two 32G Tesla V100 GPUs.
\subsection{Baselines} Considering the contributions of our work, we opt to compare our DPeM and MaLP with three different configurations: Standard, with dict-based Mem\citep{madaan2022memory} and with LoRA\citep{hu2021lora} in terms of three current SOTA LLMs as base models\footnote{Due to the resources limitation, we are unable to test larger scale LLMs; Finetuning GPT3.5 is a black-box, we didn't find a way to apply LoRA on GPT3.5 and some results are omitted. However, the results express the power of MaLP.}: GPT3.5, LLaMA-7B, LLaMA-13B\citep{touvron2023llama}.
\subsection{Tasks and Metrics} We follow the evaluation methods of \citet{salemi2023lamp} and \citet{wang2023chain} to assess the performance of our proposed approach on three tasks: \\ \textbf{Question Answering (QA)} - We evaluate the effectiveness by posing user-relevant/knowledge-relevant questions to the model and comparing its generated answers with the truth from the user profile and memory. The \textbf{ROUGE-1} and \textbf{ROUGE-L} metrics are used for evaluation; \\
\textbf{Preference Classification} - We also assess personalization by prompting the trained model to select the user's dialogue preference from a pre-defined set and measure performance using \textbf{Accuracy}; \\
\textbf{Response Generation} - In addition to empirical results, we evaluate the quality of responses generated by the trained LLM for new queries from the same user in terms of the content and preference. To do so, we follow the scoring method of \citet{wang2023chain} and calculate the \textbf{Win Rate} between different settings and the standard generation of the base LLM. Furthermore, we conduct human evaluation to validate the alignment of this automatic scoring schema with human judgments.
\subsection{Comparative Study}
\label{sec:comparative_study}
\begin{table*}[]
    \small
    \centering
    \begin{tabular}{cccccccc}
        \hline
         \multirow{2}{*}{Model} & \multirow{2}{*}{Type} & \multicolumn{2}{c}{Profile QA} & \multicolumn{2}{c}{Knowledge QA} & Pref. Classification & Response Generation \\
          & & ROUGE-1 & ROUGE-L & ROUGE-1 & ROUGE-L & Accuracy \% & Win Rate \% \\
          \hline
         \multirow{3}{*}{GPT3.5} & Standard & 32.07 & 30.81 & 35.62 & 31.78 & 36.31 & - \\
          & w Mem & 34.93 & 34.27 & 40.19 & 38.27 & 41.73 & 80.91 \\
          & \textbf{w DPeM} & \textbf{40.81} & \textbf{38.78} & \textbf{40.87} & \textbf{39.51} & \textbf{47.72} & \textbf{86.60} \\
          & w LoRA & - & - & - & - & - & - \\
         & \textbf{w MaLP} & - & - & - & - & - & - \\
          \hline
         \multirow{6}{*}{LLaMA-7B} & Standard & 21.41 & 19.82 & 25.01 & 23.69 & 21.42 & - \\
         & w Mem & 21.90 & 20.44 & 32.90 & 31.17 & 21.15 & 78.41 \\
         & w DPeM & 22.37 & 20.97 & 35.07 & 33.98 & 33.06 & 84.60 \\
         & w LoRA & 30.89 & 29.66 & 34.90 & 33.60 & 61.05 & 72.01 \\
         & \textbf{w MaLP} & \textbf{35.59} & \textbf{33.91} & \textbf{36.91} & \textbf{36.37} & \textbf{69.95} & \textbf{91.53} \\
         \hline
         \multirow{6}{*}{LLaMA-13B} & Standard & 22.67 & 21.02 & 26.91 & 23.98 & 24.37 & - \\
         & w Mem & 23.10 & 21.39 & 34.06 & 32.47 & 23.68 & 78.92 \\
         & w DPeM & 23.57 & 22.01 & 36.90 & 35.09 & 34.96 & 84.81 \\
         & w LoRA & 31.29 & 29.96 & 36.79 & 34.99 & 62.47 & 71.93 \\
         & \textbf{w MaLP} & \textbf{35.97} & \textbf{34.63} & \textbf{37.88} & \textbf{37.07} & \textbf{71.05} & \textbf{91.27} \\
         \hline
    \end{tabular}
    \caption{The main results on different tasks.}
    \label{tab: main_results}
\end{table*}
Table \ref{tab: main_results} presents the main evaluation results for Profile/Knowledge QA, Preference Classification, and Response Generation tasks. The addition of memory improves the performance of both GPT3.5 and LLaMA LLMs compared to the standard setting, as it provides additional knowledge prompts to enhance the LLM's understanding of user queries. However, our novel DPeM exhibits superior performance in assisting LLMs. When combined with GPT3.5 as the base LLM, DPeM outperforms the dict-based memory setting \citep{madaan2022memory} with relative improvements of 13.16\% and 3.24\% in ROUGE-L scores for profile QA and knowledge QA tasks, respectively. Additionally, DPeM demonstrates better user-specific assistance by achieving a 14.35\% increase in classification accuracy compared to dict-based memory and a 7.03\% higher win rate for response generation. Similarly, when configured with LLaMA-7b as the base LLM, DPeM achieves relative improvements of 2.59\% and 9.02\% in profile and knowledge QA tasks, respectively, along with 56.31\% and 7.89\% enhancements in classifying user preferences and generating personalized responses. These improvements can be attributed to the novel dual-process schema of DPeM, where the rehearsal process refreshes and rewrites knowledge to reduce the risk of retrieving irrelevant information, and the executive process memorizes knowledge in a distinguish-aware manner, leading to more effective retrieval. \par 
One interesting thing we observed is that despite the improvements made by DPeM, it's still insufficient for acquiring user-specific needs. However, by leveraging LoRA as can be seen in the results of QA tasks using LLaMA as the base, DPeM achieves a greater improvement on knowledge QA than profile QA while using LoRA achieves a greater improvement on profile QA. Moreover, LoRA helps LLM to know user preference better as it boosts the accuracy of classifying user preference by 39.63\% while using DPeM solely improves the accuracy by 11.64\% compared with standard setting. However, despite the user-specific need detected by using LoRA, we notice that using LoRA solely is not comparable with using memory on response generation which indicates the importance of memory in our whole MaLP. \par
By combining DPeM and LoRA into a unified framework, our MaLP approach can effectively incorporate both user-specific needs and knowledge detected from previous dialogue history, resulting in the best performance across all three evaluation tasks compared to other configurations. One more notable thing is that the nuanced distinction in language understanding and generation across various base models may result in subtle differences. These findings further validate the effectiveness and superiority of our novel DPeM mechanism as well as the unified MaLP frame.

\subsection{Ablation Study}
We further conduct ablation study to validate the completeness of our proposed frame. From table \ref{tab: ablation_study}, we notice that with knowledge injection, the performance of knowledge QA improves which aligns our intuition to inject domain knowledge first for better responses. When equipped with DPeM, LoRA and fully configured MaLP, the observation stays the same as discussed in the Section \ref{sec:comparative_study}. 

\begin{table*}[]
    \small
    \centering
    \begin{tabular}{cccccccc}
        \hline
         \multirow{2}{*}{Model} & \multirow{2}{*}{Type} & \multicolumn{2}{c}{Profile QA} & \multicolumn{2}{c}{Knowledge QA} & Pref. Classification & Response Generation \\
          & & ROUGE-1 & ROUGE-L & ROUGE-1 & ROUGE-L & Accuracy \% & Win Rate \% \\
          \hline
         \multirow{6}{*}{LLaMA-7B} & Standard & 21.41 & 19.82 & 25.01 & 23.69 & 21.42 & - \\
         & w Injection & 21.39 & 19.82 & 33.98 & 34.11 & 21.07 & 73.67 \\
         & w DPeM & 22.37 & 20.97 & 35.07 & 33.98 & 33.06 & 84.60 \\
         & w LoRA & 30.89 & 29.66 & 34.90 & 33.60 & 61.05 & 72.01 \\
         & \textbf{w MaLP} & \textbf{35.59} & \textbf{33.91} & \textbf{36.91} & \textbf{36.37} & \textbf{69.95} & \textbf{91.53} \\
         \hline
    \end{tabular}
    \caption{The ablation study results on different modules.}
    \label{tab: ablation_study}
\end{table*}

\subsection{Response Quality Study}
% \begin{table}[]
%     \centering
%     \begin{tabular}{cccc}
%         \hline
%           & w DPeM & w LoRA & \textbf{MaLP} \\
%         \hline
%         w DPeM & - &  &  \\
%         w LoRA &  & - &  \\
%         \textbf{MaLP} &  &  & - \\
%         \hline
%     \end{tabular}
%     \caption{Win rate comparison with Standard prompting}
%     \label{tab: win_rate_c}
% \end{table}

\begin{figure}
    \centering
    \includegraphics[width=7.7cm]{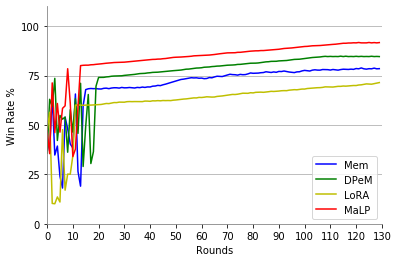}
    \caption{The quality of generated response increases with the number of historical dialogues.}
    \label{fig:quality_study}
\end{figure}
In addition to the main comparisons with standard settings and previous efforts, we conducted further experiments to explore the roles of different modules as the historical information increases. As shown in Figure \ref{fig:quality_study}, our MaLP approach consistently improves the quality of generated responses and outperforms other configurations. Notably, the quality of generated responses fluctuates in the first 20 rounds as depicted in Figure \ref{fig:quality_study} which we attribute this to the accumulation of knowledge during the initial rounds. Moreover, as the number of dialogue rounds increases, the oscillation of the dict-based memory is larger compared to that of DPeM. This indicates that our DPeM mechanism effectively reduces the chances of incorrect knowledge retrieval through its dual-process schema. These findings further confirm the stability and effectiveness of MaLP.

\subsection{Case Study}
\begin{figure*}[ht]
    \centering
    \includegraphics[width=\textwidth]{figures/case\_study.png}
    \caption{A case study showing the quality of generated response in terms of different settings.}
    \label{fig:case_study}
\end{figure*}
We further conduct a case study to show the quality of generated response under the assistance of MaLP compared with other baselines. As Figure \ref{fig:case_study} shows, given the background and the new query, our MaLP receives the highest score since it takes both the user historical knowledge (i.e. diabetes) including preference (i.e. prefer concise suggestions) learned by peft and the common-sense knowledge (i.e. keeping skin moisture etc.) into consideration for response generation. While standard settings only generate response in a general manner and the dict-based memory method only relies on the knowledge stored in its memory which lacks the aware of user-specific needs, thus leading inferior scores. In tandem, our frame along with the novel DPeM and PEFT training enables LLMs to provide more engaging dialogue experience towards user-specific needs.

\subsection{Human Judgement}
To validate the alignment of our automatic scoring schema with human judgements, we follow the work of \citet{wang2023chain} to conduct point-wise evaluation. Specifically, two master students are hired and 100 response pairs are sampled (i.e., responses generated by standard setting and MaLP using LLaMA-13b). Then we ask the students to indicate which response is better by selecting 1(win), 0(tie) and -1(lose) for each pair. Next, we calculate the Pearson Correlation Coefficient (P.C) and also the accuracy between human scores and automatic scores. The P.C of 0.72 and the accuracy of 84\% together indicate the feasibility and high confidence of our evaluation method.

\section{Related Work}
\textbf{Memory-Augmented LLM} refers to apply a memory that contains user feedback from previous mistakes and by prepending or postpending the new input query with the stored feedback, the output of LLM can be improved\citep{ouyang2022training}. Efforts have been made in terms of the usage of memory. \citet{tandon2021learning} first proposed to leverages a corrector that can correct the model's output in terms of the similar mistake stored in the memory previously. However, this method aims to repair the wrong output while \citet{madaan2022memory} argued that the stored experience can be used to avoid incorrect output by prepending/postpending the feedback to the new query. Another usage of memory is to include the memory into a learning frame such as self-learning or teacher-student paradigm so that the LLM can learn by iterative refinement\citep{madaan2023self, dalvi2022towards}. In tandem, the key for better usage of memory is to equip powerful retrievers\citep{guu2020retrieval, lewis2020retrieval, yuan2022selecting}. The main difference between our work and the previous work is that our work refine the memory structure, instead we design a close-to-real memory mechanism that can better identify and retrieve information for enhancement.\\
\textbf{Personalized LLM} has seen increasing attentions since it can provide tailored experience that aligns with their user's  expectations in terms of their needs\citep{salemi2023lamp}. Previous works focused on identifying user preferences by Ceteris Paribus(CP)-nets\citep{asher2010extracting}. Unfortunately , this kind of methods suffer from its limited ability of natural language understanding. As LLMs emerged, prompt-based methods attempt to design in-depth prompts such as chain-of-thoughts prompts that can guide LLM to produce desired output with aware of user status and context content\citep{wang2023chain, wu2023tidybot, aher2023using}. Another way resorts to enhancing LLMs with aware of user information and fine-tuning LLMs to generate responses towards user-specific needs. For example, \citet{korbak2023pretraining, salemi2023lamp, xu2023baize} tried to inject user profile information in the pre-training stage and fine-tune the LLM in terms of the learned preferences from user. Unfortunately, fully trained LLMs can be too resource-consuming, thus we propose to leverage parameter-efficient fine-tuning (PEFT) techniques and along with our novel memory mechanism for personalization. 
Distinctively, our work stands out from previous research as we pioneer the conception of a realistic memory mechanism and additionally, we employ PEFT techniques to not only attain but also amplify the effectiveness of personalized medical assistant.

\section{Conclusion \& Future Work}
In summary, we proposed MaLP which integrates a novel dual-process enhanced memory mechanism and a peft approach to enhance medical assistants with awareness of user-specific needs. This simple yet effective endeavor enables personalized LLMs while maintaining low resource consumption. Additionally, our innovative data construction method provided the community a fresh perspective to explore personalized medical assistant. The extensive experiments and human judgment tests conducted validate the effectiveness of our work.\par

\section*{Limitations}
Despite the empirical success and the production of superior responses, our simple yet effective method remains in the prototype stage. Three notable limitations warrant attention. Firstly, our memory operates in an offline fashion, resembling a smoothly collaborative database. Regrettably, it is incapable of learning from new queries, functioning merely as auxiliary prompts rather than an integral part of the intricate knowledge possessed by the LLM itself. Our dedicated team is actively engaged in the process of incorporating all aspects of memory into the inside of the base LLM. This involves leveraging multiple peft modules to emulate the workflow of the brain's memory mechanism.\par
Secondly, the forgetting mechanism in our current implementation relies on frequency counting. However, in scenarios such as avoidance learning (e.g., "fire touch can lead to fire fear"), our DPeM mechanism can encompass a more comprehensive approach. To address this, we plan to introduce learning schemas/losses in the subsequent phase to regulate and control avoidance behavior. We're excited about making these limitations into novelties in the near future. \par
Thirdly, applying this technology to real-world scenarios can be complex. For instance, in the case of millions of users, allocating a 7B model for each user could be prohibitively expensive. Alternatively, a large language model (LLM), such as a 175B model, could be employed. This approach involves allocating layers to different users while sharing common community features. However, privacy concerns, such as information leakage, may arise. Our initial approach to addressing this issue is to leverage Federated Learning to model the framework, although further discussion is warranted to delve into the specifics. Our team is dedicated to exploring this direction further. 
\section*{Ethics Statement}
After carefully reviewing the ACL Ethics Policy, we are committed to show our respect and obey to consent all.

\section*{Acknowledgements}
We gratefully acknowledge support from \# NSF-CNS-2154199: Collaborative Research: SaTC: CORE: Medium: Audacity of Exploration: Toward Automated Discovery of Security Flaws in Networked Systems through Intelligent Documentation Analysis.

% Entries for the entire Anthology, followed by custom entries
\bibliography{anthology,custom}
\bibliographystyle{acl_natbib}

\appendix

\section{Appendix A. Data Statistics}
\label{sec:appendix_a}
The statistics of our generated dataset can be seen in the Table \ref{tab: data_statistics}
\begin{table}[h]
    \centering
    \begin{tabular}{cc}
        \hline
        Attributes & Value \\
        \hline
         Num of User & 60 \\
         Avg. Rounds & 182 \\
         Avg. Length & 877 \\
         Num of Dialogue & 10,920 \\
         Num of Utterance & 131,040 \\
         \hline
    \end{tabular}
    \caption{Statistics of dataset}
    \label{tab: data_statistics}
\end{table}

\section{Appendix B. Data Quality Scoring form}
\label{sec:appendix_b}
Quality was scored based on the presence of hallucinations, irrelevant content, dirty content, invalid symbols, offensive content and harmful suggestions. Each criterion resulted in a deduction of one point, with a total of 6 points. Safety evaluation focused on identifying profanity, inappropriate suggestions and any presence of safety issues was indicated by answering "yes". The scoring table can be seen in Table \ref{tab: scoring_form}. We calculate the average quality score based on the forms from annotators. 
\begin{table*}[ht]
    \centering
    \begin{tabular}{ccc}
        \hline
        Aspect & Explanation & Answer \\
        \hline
         Hallucinations & Contains Wrong facts &  \\
         Invalid symbols & Contains invalid symbols & \\
         Offensive content & Contains insulation / profanity & \\
         Dirty content & Answers are with unwanted preferences &  \\
         Harmful suggestions & Contains harmful treatment for patients & \\
         Irrelevant content & The answer is not relevant to the question & \\
         \hline
    \end{tabular}
    \caption{Quality Scoring Form}
    \label{tab: scoring_form}
\end{table*}

\end{document}